\documentclass{article}

\PassOptionsToPackage{numbers, compress}{natbib}
\usepackage[final]{neurips_2024}

\usepackage[utf8]{inputenc} 
\usepackage[T1]{fontenc}    
\usepackage{hyperref}       
\usepackage{url}            
\usepackage{booktabs}       
\usepackage{amsfonts}       
\usepackage{nicefrac}       
\usepackage{microtype}      
\usepackage{xcolor}         
\usepackage{graphicx}       
\usepackage{amsmath}       

\title{DMAGT: Unveiling miRNA-Drug Associations by Integrating SMILES and RNA Sequence Structures through Graph Transformer Models}

\author{%
  Ziqi~Zhang \\
  Department of Computer Science \\
  Brown University \\
  Providence, RI 02912, USA \\
  \texttt{ziqi\_zhang2@brown.edu}
}

\begin{document}

\maketitle

\begin{abstract}
MiRNAs, due to their role in gene regulation, have paved a new pathway for pharmacology, focusing on drug development that targets miRNAs. However, traditional wet lab experiments are limited by efficiency and cost constraints, making it difficult to extensively explore potential associations between developed drugs and target miRNAs. Therefore, we have designed a novel machine learning model based on a multi-layer transformer-based graph neural network, DMAGT, specifically for predicting associations between drugs and miRNAs. This model transforms drug-miRNA associations into graphs, employs Word2Vec for embedding features of drug molecular structures and miRNA base structures, and leverages a graph transformer model to learn from embedded features and relational structures, ultimately predicting associations between drugs and miRNAs. To evaluate DMAGT, we tested its performance on three datasets composed of drug-miRNA associations: ncDR, RNAInter, and SM2miR, achieving up to AUC of 95.24±0.05. DMAGT demonstrated superior performance in comparative experiments tackling similar challenges. To validate its practical efficacy, we specifically focused on two drugs, namely 5-Fluorouracil and Oxaliplatin. Of the 20 potential drug-miRNA associations identified as the most likely, 14 were successfully validated. The above experiments demonstrate that DMAGT has an excellent performance and stability in predicting drug-miRNA associations, providing a new shortcut for miRNA drug development.
\end{abstract}

\section{Introduction}

MicroRNAs (miRNAs) are small non-coding RNAs that have garnered significant attention in the field of pharmacology due to their critical role in gene regulation and their associations with a range of diseases \cite{RF1,RF2}. Studies have demonstrated that miRNAs exert influence on various cancers \cite{RF3}, diabetes \cite{RF4}, and cardiovascular diseases \cite{RF5}, particularly at the genetic level. Recently, a novel category of drugs has been developed with the capability to modulate miRNA activity, offering a promising approach for disease treatments by targeting specific miRNAs. An illustrative example of these therapeutic advancements on the application of miRNA therapeutics in cardiac conditions. Bernardo's study highlights drugs that can inhibit or diminish the expression of pertinent miRNAs, thereby mitigating their upregulation in response to cardiac stress \cite{RF6}. Moreover, miRNAs can also affect drug metabolism and toxicity \cite{RF7}. In the realm of colorectal cancer, the study conducted by Pouya elucidates that several miRNAs play a role in drug resistance by regulating genes associated with this resistance. These miRNAs, linked to drug resistance, possess significant potential as biomarkers for predicting drug responses or as targets for tailored treatments in individuals with colorectal cancer \cite{RF8}. Consequently, investigating and understanding the interplay and mechanisms between miRNAs and drugs opens a novel avenue for disease treatment.

Nonetheless, it is essential to acknowledge the difficulties and challenges inherent in exploring the relationships between miRNAs and drugs. The conventional method of wet-lab experiments, typically employed for such investigations, is often time-consuming and labor-intensive. This is particularly true given the extensive number of potential associations among a vast array of miRNAs and their numerous targets \cite{RF9}. One complicating factor is the capacity of miRNAs to regulate multiple targets concurrently, necessitating the execution of numerous experiments to construct a comprehensive map of drug-miRNA associations \cite{RF10}. Furthermore, the subtle phenotypic alterations induced by miRNA modulation can make experimental detection arduous, thereby increasing the complexity of the experiments \cite{RF11}. Financial constraints frequently impinge upon the scale and breadth of wet-lab experiments, particularly when examining a wide range of potential drug-miRNA associations. Considering these intricacies, it is imperative to innovate and adopt novel approaches to enhance the efficiency of wet-lab experiments and curtail associated costs.

Computational methods have become increasingly prevalent in supplementing and guiding wet lab experiments, offering advantages in terms of speed and cost-effectiveness compared to traditional experimental techniques. For instance, Li et al. developed a network-based model that integrates drugs, miRNAs, and genes into a heterogeneous network. This model aims to identify target miRNA-drug associations pertinent to cancer treatment \cite{RF12}. Another significant approach is matrix factorization, as exemplified by Jamali et al., who constructed a miRNA-drug similarity matrix to facilitate the prediction of microRNA–drug associations \cite{RF13}. With the advent and proliferation of deep learning, strategies that combine network-based models and deep learning have gained traction. A notable example is the deep learning model based on Convolutional Neural Networks, designed by Deepthi et al., to predict associations between miRNAs and drug resistance \cite{RF14}. These computational strategies contribute significantly to a more efficient elucidation of miRNA-drug relationships.

However, the methods mentioned above have limitations, particularly in fully exploiting the feature and positional information of drugs and miRNAs. To address this, our paper introduces a novel graph-based deep learning model that incorporates a multi-layer transformer-based graph neural network, which we have named DMAGT, to predict associations between miRNAs and drugs. We encode the chemical information of miRNAs and drugs into vector formats and construct a graph where each miRNA or drug is represented as a node with its chemical features, and each interaction as an edge, thereby utilizing DMAGT to engage in edge prediction tasks for predicting miRNA-drug associations. The performance of our model is evaluated through five-fold cross-validation across three distinct databases. Additionally, to further substantiate the efficacy of our model, we select specific miRNAs and drugs for case studies, conducting cross-validation between databases.

\section{MATERIALS}

\subsection{Datasets}

Presently, a substantial array of databases is dedicated to miRNAs, drugs, and miRNA-drug associations. We opted to utilize three databases specifically curated for miRNA-drug associations, namely ncDR, RNAInter, and SM2miR \cite{RF15,RF16,RF17}. Through these databases, we aggregated a total of 12,323 miRNA-drug associations. Based on these associations, we systematically extracted 516 unique drugs and 2,213 miRNAs, each annotated with their respective CIDs and identifiers.

Considering the lack of chemical information for miRNAs and drugs in the initial databases, we expanded our data acquisition by integrating two supplementary databases to procure more detailed information about miRNAs and drugs. For miRNAs, we sought the nucleotide sequences essential for feature construction, leading us to utilize the miRBase database \cite{RF18}. This platform facilitates the extraction of requisite nucleotide sequences via miRNA identifiers. In the context of drugs, we required detailed molecular structures that would serve as features. Therefore, we accessed the PubChem database, renowned for its comprehensive compound datasets \cite{RF19}. Using the drug CIDs, we retrieved the corresponding SMILES from PubChem, which express molecular structures in string notation \cite{RF20}. SMILES is not only informative in detailing the drug's chemical constitution but also optimize the ensuing embedding procedures.

After this data consolidation, we assembled a comprehensive dataset encompassing miRNA-drug associations, supplemented by the pertinent chemical information. The next phase involves the embedding of this chemical data, transforming it into vector representations to be utilized as features in our models.

\begin{table}
  \caption{The statistical analysis of miRNAs, drugs, and their associations across three datasets}
  \label{table1}
  \begin{tabular}{lllll}
    \hline
    Databases   & ncDR & RNAInter & SM2miR & Total \\
    \hline
    Drugs   & 95 & 283 & 138 & 516   \\
    miRNAs  & 624 &	1009 &	580 &	2213 \\
    Associations  & 4457 &	5740 &	2126 &	12323 \\
    \hline
  \end{tabular}
\end{table}

\subsection{Representations of Chemical Information of miRNAs and Drugs}

Chemical information, such as SMILES and nucleotide sequences, can accurately depict the characteristics of miRNAs and drugs in practical applications. However, this information is inherently discrete and cannot be directly integrated into computational models. Consequently, we require a methodology to vectorize this information, enabling its involvement in computational calculations within our model—a process we refer to as "word embedding."

The conventional approach for vector representation of words is the One-Hot Representation. This technique requires a dictionary of length N, generating an N-dimensional array for each word, where the position representing the word is marked as 1, and all other positions are set to 0. Despite its utility in representing discrete information in a vector space, this method has two primary deficiencies when characterizing miRNAs and drugs. First, the resultant vector is overly sparse, laden with an excessive presence of zeros, particularly in representing miRNAs. Depending on the database, the SMILES of our miRNAs contain up to 61 distinct characters, necessitating the generation of a vector with 60 zeros and one 1 for each SMILES. Such sparsity is memory-intensive during model training and contributes to reduced computational efficiency. Secondly, this representation fails to capture the semantics of each atom within the SMILES. For instance, "C=C" signifies a double bond between two carbon atoms, but One-Hot Representation can only differentiate the symbol “C” and “=”, losing the semantics of both the carbon atoms and the double bond, which impairs the informational richness of the features, thereby potentially diminishing model performance.

In the realm of Natural Language Processing (NLP), a more prevalent strategy is the employment of Word2Vec \cite{RF21,RF22}, a model that not only converts words into dense vector representations but also preserves their semantic relationships. Although Word2Vec is predominantly utilized for embedding in natural language contexts, its applications have extended to biological and chemical domains \cite{RF23}. We adopted the Continuous Bag of Words (CBOW) algorithm, one of the two principal training algorithms offered by Word2Vec. Throughout the training, the vector for each word undergoes continual computation, ultimately yielding a corresponding vector for every word in the dictionary.  In the study, SMILES strings of drug molecules and nucleotides are embedded into 64-dimensional vectors, with each vector representing the semantics of a single atom or nucleotide in the low dimensional space. To maintain uniform feature lengths during training, we set the sentence length threshold at 128, employing 0 padding for sequences with fewer components and applying truncation for those exceeding this limit. Through this process, we derived representations of miRNAs and drugs that are readily integrable into computational models.

\subsection{Representations of Associations between miRNAs and Drugs}

Graph structures represent a trending data configuration in contemporary deep learning research, finding extensive applications across various domains including medicine, social networking, and e-commerce \cite{RF24,RF25,RF26}. Given that the objective of this study is to investigate and predict potential associations between various miRNAs and drugs, we can abstract this problem into a graph problem comprising two distinct node types: miRNAs and drugs. If an association exists between a miRNA and a drug, an edge is established between the two nodes; otherwise, no edge is present. Furthermore, we utilize the sequences of SMILES for miRNAs and nucleotide sequences for drugs, processed through word embedding, as node attributes.

\section{METHODS}

\subsection{Problem Definition}

This study is dedicated to identifying potential unknown associations between miRNAs and drugs, building upon the known associations between them. We expect to predict unknown associations by learning the patterns of known miRNAs and drugs, which consist of a graph composed of the known miRNAs and drugs and the information contained within the nodes. By employing the word embedding algorithm and graph data structure, we transform the problem into one that involves a graph composed of two types of nodes: miRNAs and drugs. In this graph, the edges represent the associations between miRNAs and drugs. Such a network-based approach enhances the visualization of the associations between miRNAs and drugs, and simultaneously aids in identifying complex patterns and relationships that might not be as evident in tabular or other data formats. The problem we are investigating can be abstracted as a link prediction problem. Link prediction is a fundamental task in the field of network science, focusing on estimating the likelihood of the existence of links between nodes in a network. In the context of our research, this translates to predicting whether there exists an edge between any given miRNA node and any given drug node within the network constituted by miRNAs and drugs.

\begin{figure*}[h]
    \centering
    \includegraphics[width=6in]{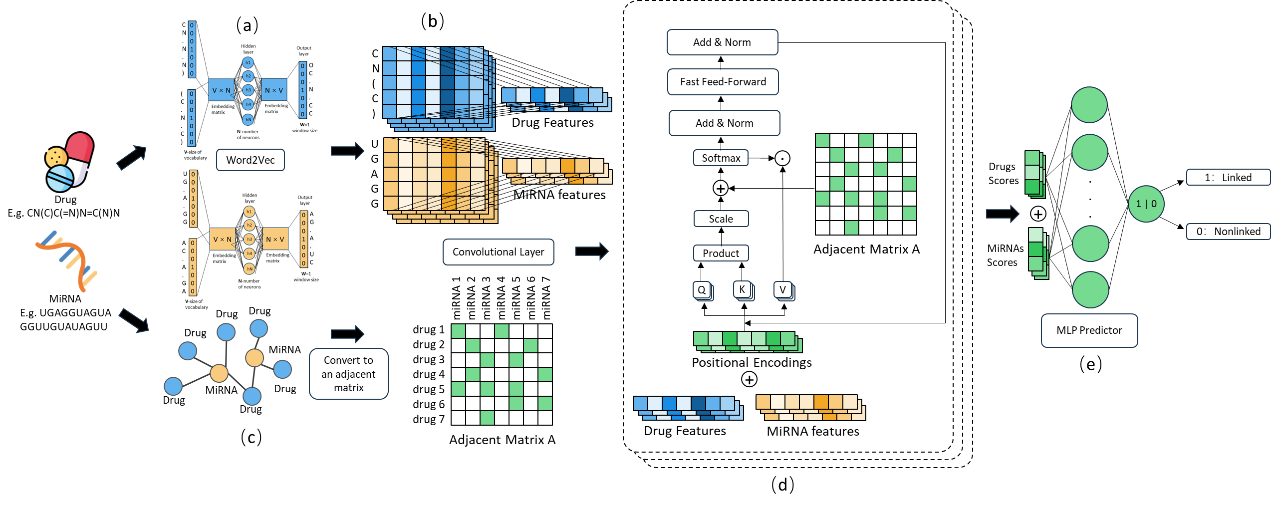}
    \caption{Overview of DMAGT. a) Embedding chemical information of miRNAs and drugs into low-dimensional vectors with Word2Vec. b) Reducing the number of dimensions of the miRNAs and drugs vectors to 1-dimensional vector through a convolutional layer c) Constructing an adjacent matrix according to the known associations between miRNAs and drugs. d) Learning the chemical information and associations with a multi-layer transformer-based graph neural network to compute node scores for drugs and miRNAs. e) Predict the unknown associations with node scores.}
    \label{figure1}
\end{figure*}

We achieve link prediction through a two-tiered model structure. This model comprises a graph transformer model and an MLP (Multilayer Perceptron) predictor component.

\subsection{Multi-layer Transformer-based Graph Neural Network}

Transformers have become a significant model in the field of Natural Language Processing (NLP), making substantial contributions in the domains of language translation, text generation, and conversational AI \cite{RF27}. A crucial innovation of this model is the self-attention mechanism. This mechanism enables the model to score each token in the input sequences, reflecting the importance of that “token” to other tokens. If we consider each node in Graphs as a token, we can migrate this mechanism from the NLP domain to the graph neural network domain.

Dwivedi et al. proposed a novel architecture of transformer that is adapted for graphs \cite{RF28}. This architecture introduces two primary innovative measures that make the transformer model seamlessly compatible with Graphs. In the graph transformer model, we use Laplacian eigenvectors as the positional encoding (PE). The formula is as follows:
\begin{equation}
  \Delta = I - D^{-\frac{1}{2}} A D^{-\frac{1}{2}} = U^T \Lambda U
\end{equation}
where A denotes the adjacency matrix corresponding to the graph, D is the degree matrix, and AU represents eigenvalues and eigenvectors. For each node i, we have the k smallest non-trivial eigenvectors as its positional encoding. Laplacian Positional Encodings (PEs) expand upon the PE used in the original transformers by Vaswani et al. \cite{RF27}, adapting them for graphs. This enhancement effectively captures distance-aware information, ensuring that nodes in proximity have similar positional features, while those further apart exhibit dissimilar positional features \cite{RF29}. The second innovation involves performing attention on each node’s connectable neighborhood \cite{RF28}. Unlike the original transformer, where the importance of each token is calculated relative to other tokens, the graph transformer model leverages the sparsity of nodes in graphs, meaning each node has clear edges leading to other nodes. This approach avoids the need to generate a fully connected graph for computation, significantly optimizing the model's computational efficiency \cite{RF27,RF28}.

Based on the concepts, we constructed a multi-layer transformer-based graph neural network with the following architecture. We re-extract information from the embeddings of miRNAs and drugs through a convolutional layer, learning the interaction information within sequences of words and reducing the original embeddings to one-dimensional vectors with a length equal to the hidden size. Through our testing, we found that using a kernel size equal to the length of the sequence in the embeddings yields the best results. Therefore, we set the kernel size to 128, with the in-channel equal to the length of the sequence and the out-channel equal to the length of the hidden size d. This process can be represented as:
\begin{equation}
    h^0 = \text{ReLU}(W \cdot X)
\end{equation}
where  ${h}^{0} \in {\mathbb{R}}^{n\times d}$ denotes the initial hidden features, ${W}$ denotes the convolution kernel, $ X \in {\mathbb{R}}^{{d}_{f}\times {d}_{s}} $ denotes the input node features, ${n}$ denotes the number of sequences, s denotes the sequence length, ${f}$ denotes the dimension of words, ${*}$ represents a convolution operation. We also perform a linear transformation on the positional encoding (PE), converting its length to equal the hidden size. Subsequently, we add the positional encoding to the hidden features obtained by the convolution operation. At this stage, we have:
\begin{equation}
  {\hat{h}}^{0} = {h}^{0} + W\Delta
\end{equation}
where $W \in \mathbb{R}^{d \times d_{\Delta}}$, $\delta \in \mathbb{R}_{\Delta}$ denotes the positional encoding. For the second step, for each hidden layer, we incorporate a multi-head self-attention module, which uses the self-attention matrices $Q^k, K^k, V^k \in \mathbb{R}^{d \times d_k}$, and adjacent matrix $A \in \mathbb{R}^{d \times d}$ to calculate the scores of nodes, and these are then passed on to the next layer:
\begin{equation}
    \text{Attn} = \frac{(\widehat{h}^l Q^{k,l}) \cdot (\widehat{h}^l K^{k,l})^T}{\sqrt{d_k}} \odot A
\end{equation}
\begin{equation}
    \text{head}_k = \text{SoftMax}(\text{Attn}) (\widehat{h}^l V^{k,l})
\end{equation}
\begin{equation}
    \widehat{h}^{l+1} = O^l \left( \text{Concat}(\text{head}_1, \ldots, \text{head}_n) \right)
\end{equation}
where $O^l \in \mathbb{R}^{d \times d}$, $k$ denotes the number of attention heads, $l$ denotes the number of layers, $d_k$ denotes the number of head dimensions, and $\odot$ represents a Hadamard product operation. Afterward, we integrate a feed-forward network, adding non-linear components through activation functions to perform further feature extraction. Additionally, we incorporate residual connections and the batch normalization mechanism:

\begin{equation}
    \dot{h}^{l+1} = \text{Norm}_1(\widehat{h}^l + \widehat{h}^{l+1})
\end{equation}
\begin{equation}
    \ddot{h}^{l+1} = W_2 \, \text{ReLU}(W_1 \dot{h}^{l+1})
\end{equation}
\begin{equation}
    \widehat{h}^{l+1} = \text{Norm}_2(\dot{h}^{l+1} + \ddot{h}^{l+1})
\end{equation}
where ${W}_{1} \in {\mathbb{R}}^{2d\times d}$  , ${W}_{2} \in {\mathbb{R}}^{d\times 2d}$, and ${Norm}_{1}$ and ${Norm}_{2}$ denote batch normalization operations. Having acquired the hidden features for each node, we then proceed by concatenating the hidden features of two linked nodes. These concatenated features are then fed into a three-layer multilayer perceptron for link prediction to compute the scores of the edge connecting the pair of nodes. Subsequently, the SoftMax function is employed to convert all the scores into probabilistic representations. The methodology for this process is delineated as follows:

\begin{equation}
    h_{ij} = \text{Concat}(h_i, h_j)
\end{equation}
\begin{equation}
    s_{ij} = W_3 \left( \text{ReLU} \left( W_2 \left( \text{ReLU} \left( W_1 (h_{ij}) \right) \right) \right) \right)
\end{equation}
\begin{equation}
    p_{ij} = \text{SoftMax}(s_{ij})
\end{equation}
where ${p}_{ij}$ denotes the probability of the presence of an edge between node i and node j, which implies the likelihood of an association between drug i and miRNA j.

\section{EXPERIMENTS}

\subsection{Experimental Setup}

For analysis, we divided the samples into training and testing datasets. The training dataset constituted 80\% of the sample associations as positive samples, supplemented by the same number of artificially generated non-interacted drug-miRNA pairs as negative samples. Conversely, the testing dataset was composed of the remaining 20\% of the sample associations as positive samples, and the equivalent amount of non-interacted pairs as negative samples.

Our model was trained using the training graph, and its performance was evaluated using the testing graph. In our model, we employed the Non-dominated Sorting Genetic Algorithm II (NSGA-II) \cite{RF30}, utilizing the Optuna \cite{RF31} package for hyperparameter tuning. The Adam optimization algorithm was chosen, with a learning rate set at 0.00004, weight decay at 0.0033, and 256 hidden layers. To rigorously test our model, we implemented 5-fold cross-validation, running 150 epochs for each training iteration. We conducted the experiment on a device equipped with 16GB of memory, an 11th Gen Inteli7-11800H CPU, and an NVIDIA RTX 3060 GPU.

\subsection{Experimental Results}

In this experiment, we employed a 5-fold cross-validation method to evaluate the performance of our model. As shown in Table~\ref{table2}, we tested the Acc, Sen, Spec, Prec, Mcc, and AUC values of the model in each validation. We also calculated the average and standard deviation of these values from the 5 validations to further assess the model's performance. The most outstanding performance was observed when the model was run on the ncDR dataset, where it ranked first in all metrics. When processing the RNAInter dataset, the model's performance was almost the same as in the ncDR dataset, with the average AUC differing only by 0.82\%. However, in the SM2miR dataset, the model's performance was weaker than in the previous two datasets. This is because the number of drugs, miRNAs, and associations in the SM2miR dataset is only half that of ncDR and RNAInter. Therefore, this part of the performance can be attributed to the decline in performance due to the insufficient training set size. Nevertheless, even under these circumstances, the model's AUC still maintained around 84.96\%. Additionally, for visualization purposes, we plotted the AUC and AUPR of the model in Fig (3) – Fig (5). We can observe that in the validation results of the three datasets, the ROC curves lean towards the top-left corner, and the PR curves lean towards the top-right corner, maximizing the area under the curve. This visually demonstrates the excellent AUC and AUPR values of our model.
\begin{table}
  \caption{The performance of the proposed method in different datasets.}
  \label{table2}
  \begin{tabular}{lllllll}
    \hline
    Dataset & Fold  & Acc. (\%) &	Sen. (\%) &	Spec. (\%) & Prec. (\%) &	MCC. (\%)  \\
    \hline
    ncDR & 1 &	88.04 &	87.89 &	88.19 &	88.16 &	76.08 \\
         & 2 &	87.97 &	88.86 &	87.07 &	87.3 &	75.95 \\
         & 3 &	88.00 &	87.52 &	88.49 &	88.38 &	76.01 \\
         & 4 &	88.15 &	87.74 &	88.57 &	88.47 &	76.31 \\
         & 5 &	88.00 &	87.22 &	88.79 &	88.61 &	76.02 \\
         & Average &	88.03±0.07 &	87.85±0.62 &	88.22±0.68 &	88.18±0.52 &	76.07±0.14 \\
    \hline
    RNAInter & 1 &	87.24 &	88.59 &	85.89 &	86.26 &	74.50 \\
             & 2 &	87.02 &	89.02 &	85.02 &	85.59 &	74.10 \\
             & 3 &	87.15 &	88.85 &	85.45 &	85.93 &	74.35 \\
             & 4 &	87.15 &  88.94 & 85.37 & 85.87 & 74.35 \\
             & 5 &	87.20 &	88.59 &	85.80 &	86.19 &	74.42 \\
             & Average &	87.15±0.08 & 88.80±0.20 &	85.51±0.35 & 85.97±0.27 & 74.34±0.15 \\
    \hline
    SM2miR & 1 &	78.76 &	80.56 &	76.96 &	77.76 &	57.56 \\
           & 2 &	78.61 &	78.06 &	79.15 &	78.92 &	57.21 \\
           & 3 &	77.51 &	77.43 &	77.59 &	77.55 &	55.02 \\
           & 4 &	77.66 &	79.78 &	75.55 &	76.54 &	55.38 \\
           & 5 &	77.27 &	75.71 &	78.84 &	78.16 &	54.57 \\
           & Average & 77.96±0.68 &	78.31±1.92 & 77.62±1.46 & 77.79±0.87 & 55.95±1.35 \\
    \hline
  \end{tabular}
\end{table}

\begin{figure*}
    \centering
    \includegraphics[width=6in]{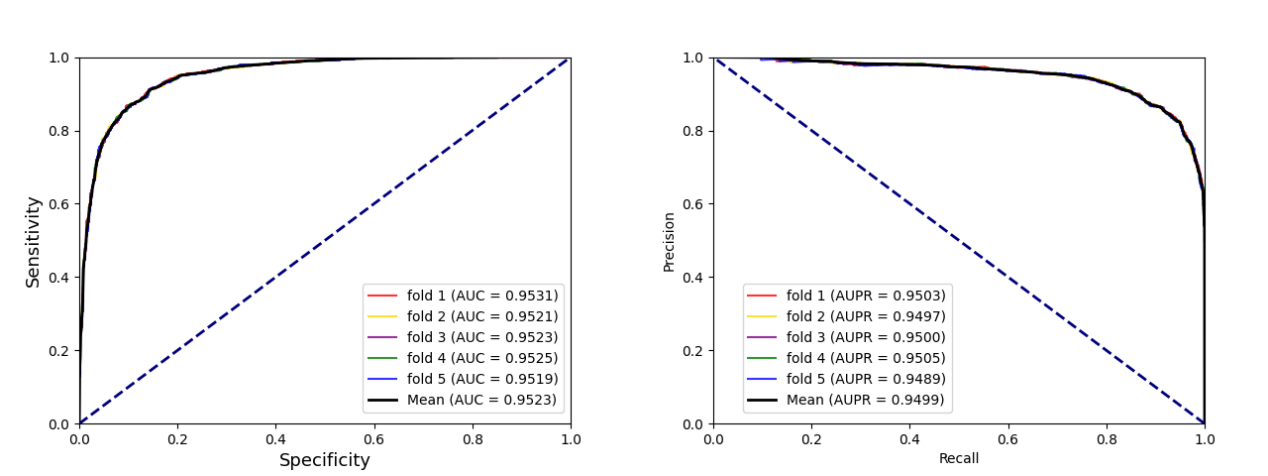}
    \caption{(a) The ROC curves of the result on the ncDR datasets. (b) The PR curves of the result on the ncDR datasets.}
    \label{figure2}
\end{figure*}
\begin{figure*}
    \centering
    \includegraphics[width=6in]{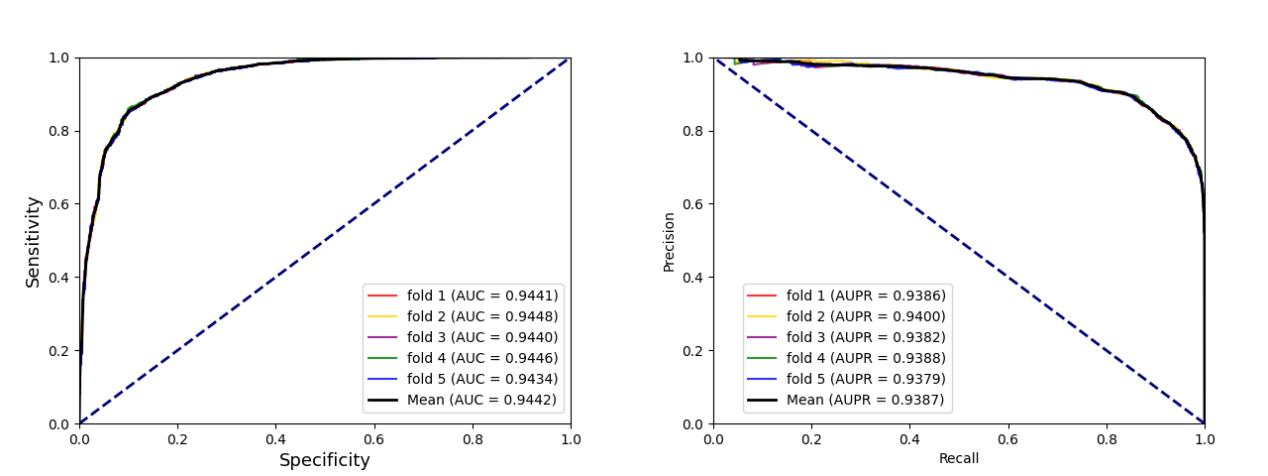}
    \caption{(a) The ROC curves of the result on the RNAInter datasets. (b) The PR curves of the result on the RNAInter datasets.}
    \label{figure3}
\end{figure*}
\begin{figure*}
    \centering
    \includegraphics[width=6in]{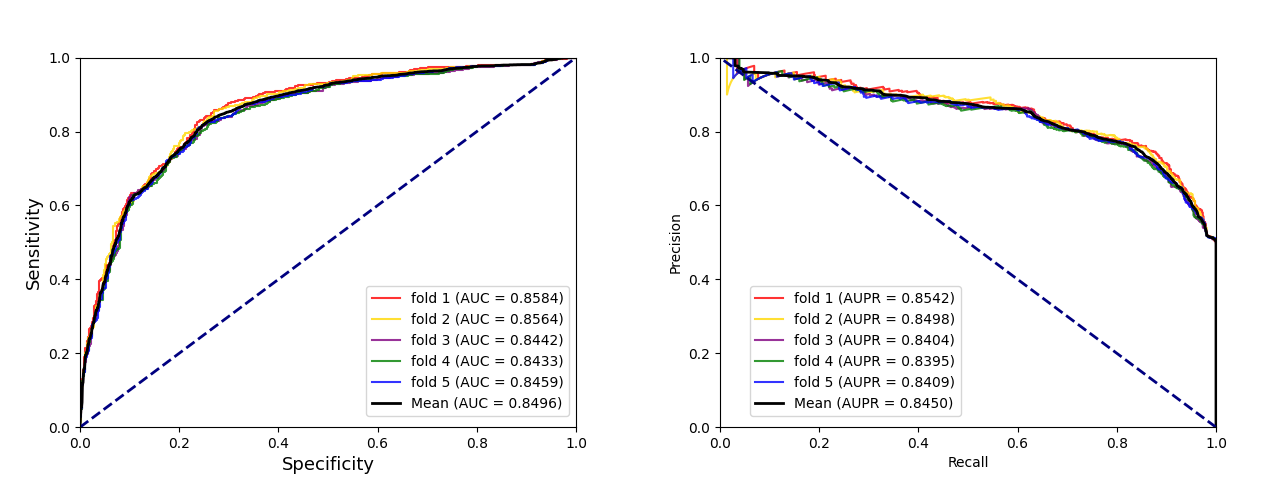}
    \caption{(a) The ROC curves of the result on the SM2miR datasets. (b) The PR curves of the result on the SM2miR datasets.}
    \label{figure4}
\end{figure*}

\subsection{Performance Comparison}

We compared our model with eight different models. The SVD model directly utilized the adjacency matrix of drugs and miRNAs for prediction \cite{RF32}. Drug-based CF, miRNA-based CF, and Neighbor-based CF employed the collaborative filtering method for association prediction \cite{RF33}. The GCMDR, EPLMI, and MDIPA methods all utilized the similarity matrix of miRNAs and drugs for model training \cite{RF13,RF34,RF35}. MFIDMA separately embedded the spatial attributes and chemical properties of miRNAs and drugs, and then used a combination of CNN and DNN for prediction \cite{RF36}. As shown in Table~\ref{table3}, after performing five-fold validation, our model surpassed all the methods with an AUC performance of 0.9525 ± 0.0006, further verifying the superior performance of this model.

\begin{table}
  \caption{Comparison of the prediction performance based on ncDR datasets.}
  \label{table3}
  \begin{tabular}{llllll}
    \hline
    Method &	AUC  \\
    \hline
    SVD-based MF \cite{RF32} & 0.6007 ± 0.0052   \\
    Drug-based CF \cite{RF33}  & 0.7313 ± 0.0008   \\
    miRNA-based CF \cite{RF33}  & 0.8235 ± 0.0015   \\
    Neighbor-based CF \cite{RF33} & 0.8644 ± 0.0009   \\
    GCMDR \cite{RF35} & 0.9359 ± 0.0006   \\
    EPLMI \cite{RF34} & 0.8971 ± 0.0009   \\   
    MDIPA \cite{RF13} & 0.9081 ± 0.0038      \\
    MFIDMA \cite{RF36} & 0.9407 ± 0.0019   \\
    Proposed Method & 0.9525 ± 0.0006   \\
    \hline
  \end{tabular}
\end{table}

\subsection{Examination of the effectiveness of the features}

\begin{table}
  \caption{The performance of the proposed method with different node features.}
  \label{table4}
  \begin{tabular}{llllll}
    \hline
    Models & DMAGT -nemb  & 	DMAGT -npe &	DMAGT  \\
    \hline
    AUC & 	70.17\% & 	95.13\% & 	95.24\%  \\
    AUPR & 	75.47\% & 	94.93\% & 	94.99\%  \\
    \hline
  \end{tabular}
\end{table}

In this study, we exam the effectiveness of the features on the ncDR dataset. Initially, we chose to remove the embedding features of the nodes in the network, retaining only the Positional Encoding for training the model. The remaining parameters were consistent with the settings mentioned in the Experiment Setup. According to Table~\ref{table4}, we observed that both AUPR and AUC experienced significant drops, with AUPR decreasing by an average of 23.51\% and AUC by 26.21\%. This indicates that the model effectively utilized the information from the embedding features, resulting in significant improvements in prediction accuracy. In another set of experiments, we removed the positional encoding, and the model's AUC and AUPR also saw a minor decline. Compared to the complete DMAGT model, the performance was slightly lower but still very close. This shows that although positional encoding contributes to the model's performance, its absence does not impact performance as much as the absence of embedding features. In contrast, the complete DMAGT model, which combines node embedding features and positional encoding, has an AUC of 95.24\% and an AUPR of 94.99\%. Among the three models, this model performed the best. Overall, the embedding features played a crucial role in the model's performance, as evidenced by the significant performance drop in the DMAGT-nemb model. Positional encoding also contributed to the model's overall performance, but its impact was not as pronounced as that of the embedding features. The complete DMAGT model, which combines both features, provided the best results.

\subsection{Case Study}

\begin{table}
  \caption{The top 10 predicted miRNAs interacting with Fluorouracil and Oxaliplatin.}
  \label{table5}
  \begin{tabular}{lll|lll}
    \hline
    \multicolumn{3}{c|}{Fluorouracil} & \multicolumn{3}{c}{Oxaliplatin} \\
    \hline
    miRNA &	Result &	Evidence & miRNA & Result & Evidence \\
    \hline
    hsa-miR-26a-5p & 	confirmed & 	RNAInter & hsa-miR-21-5p & 	confirmed & 	RNAInter \\
    hsa-miR-4758-3p & 	unconfirmed & 	unconfirmed & hsa-miR-194-5p & 	unconfirmed & 	unconfirmed \\
    hsa-miR-614 & 	confirmed & 	RNAInter & hsa-miR-15a-5p & 	confirmed & 	RNAInter \\
    hsa-let-7i-5p & 	confirmed & 	RNAInter & hsa-miR-15b-5p & 	confirmed & 	RNAInter \\
    hsa-miR-302b-3p & 	confirmed & 	RNAInter & hsa-miR-26a-5p & 	confirmed & 	RNAInter \\
    hsa-miR-32-5p & 	confirmed & 	RNAInter & hsa-miR-10b-5p & 	confirmed & 	RNAInter \\
    hsa-miR-122-5p & 	confirmed & 	RNAInter & hsa-miR-1291 & 	unconfirmed & 	unconfirmed \\
    hsa-miR-302a-3p & 	confirmed & 	PMID: 25526515 [40] & hsa-miR-549a-3p & 	unconfirmed & 	unconfirmed \\
    hsa-miR-211-5p & 	confirmed & 	PMID: 29970910 [40] & hsa-miR-183-5p & 	unconfirmed & 	unconfirmed \\
    hsa-miR-302c-3p & 	unconfirmed & 	unconfirmed & hsa-miR-155-5p & 	confirmed & 	RNAInter \\
    \hline
  \end{tabular}
\end{table}

To test the practical application of our model, we selected two drugs, 5-Fluorouracil and Oxaliplatin, for testing in the SM2miR dataset. During the tests, we excluded the known associations of these two drugs with miRNAs from the dataset; the remaining associations were used as positive samples. We also randomly selected an equal number of negative samples from unrelated drugs and miRNAs. Then, we input both positive and negative samples into our model for training. The trained model was then used to predict potential associations between these two drugs and all other miRNAs. Finally, we ranked the results by the likelihood of association from highest to lowest and validated the predicted associations in the RNAInter dataset and literature.In Table~\ref{table5}, we identified 10 miRNAs that may be associated with 5-Fluorouracil, of which 8 could be confirmed. According to the study by Deng et al., hsa-miR-302b-3p can enhance the sensitivity of hepatocellular carcinoma cell lines to 5-Fluorouracil~\cite{RF37}. Hou et al.'s study suggested that hsa-miR-302a-3p is involved in the autophagy process during 5-Fluorouracil treatment and has the potential to be used in 5-FU-based chemotherapy for colorectal cancer~\cite{RF38}. The study by Zhang et al. showed that the miR-211-5p axis promotes resistance of breast cancer cells to 5-Fluorouracil (5-FU)~\cite{RF39}.Additionally, according to Table~\ref{table5}, we identified 10 associations between Oxaliplatin and miRNAs, of which 6 could be confirmed. We consulted the literature for some of these associations. The study by Despotovic et al. indicated that Oxaliplatin altered the expression of hsa-miR-21-5p in SW620 cells related to TGF-$\beta$ signaling~\cite{RF40}. Cai et al. proposed that inhibition of the miR-15a-5p pathway enhanced the chemotherapy sensitivity of HCT116 cells resistant to Oxaliplatin~\cite{RF41}.The results show that the model effectively predicted the associations between drugs and miRNAs, and for 5-Fluorouracil, it even predicted associations not yet recorded in RNAInter, which were later confirmed in subsequent research. This reflects the accuracy and stability of our model, which can provide meaningful references and guidance for wet lab experiments.

\section{CONCLUSION}

This study provides a method for predicting the associations between drugs and miRNAs based on a multi-layer transformer-based graph neural network. This method makes full use of the chemical properties of drugs and miRNAs as well as known associations. On one hand, we employed the word2Vec algorithm to perform feature embedding on the SMILE expressions of drugs and the base expressions of miRNAs. On the other hand, we used known associations to construct graph networks of drugs and miRNAs. Based on these, we used the graph transformer model to predict unknown associations between drugs and miRNAs. Finally, we assessed the performance of the model in three different databases using five-fold cross-validation. The results are presented in the form of value matrices for Acc, Sen, Spec, Prec, Mcc, and AUC, as well as images of ROC Curves and Precision-Recall Curves, achieving the highest results with Acc=88.03±0.07, Sen=87.85±0.62, Spec=88.22±0.68, Prec=88.18±0.52, Mcc=76.07±0.14, AUC=95.24±0.05, AUPR=0.9499. In comparative experiments, our model achieved an excellent performance with AUC=95.24±0.05, surpassing all other models used for comparison. Additionally, in case studies, our model successfully predicted the associations between two different drugs and various miRNAs, with 14 out of 20 drug-miRNA associations being verifiable, further demonstrating the model's stability and accuracy. Subsequent research will focus on exploring the heterogeneous characteristics of drugs and miRNAs to further improve the network's performance.


\bibliographystyle{ieeetr}
\bibliography{Mybib}

\end{document}